\documentclass[letterpaper,10pt,journal,twoside]{IEEEtran}
\usepackage{amsmath,amsfonts,amssymb}
\usepackage{array}
\usepackage{graphicx}
\usepackage{booktabs}
\usepackage{pifont}
\usepackage{url}
\usepackage{multirow}
\usepackage{cite}
\usepackage{colortbl}
\usepackage{xcolor}
\newif\ifrevisionhighlight
\revisionhighlightfalse
\newcommand{\revision}[1]{\ifrevisionhighlight\textcolor{red}{#1}\else#1\fi}

\begin{document}

\title{DA-Nav: Direction-Aware City-Scale Vision-Language Navigation}

\author{
Ye Yuan$^{1,4,*}$,
Kehan Chen$^{3,*}$,
Xinqiang Yu$^{3,*}$,
Wentao Xu$^{1}$,
Heng Wang$^{4}$,
Libo Huang$^{2}$,
Chuanguang Yang$^{2}$,
Yan Huang$^{3}$,
Jiawei He$^{4,\dagger}$,
and Zhulin An$^{2,\dagger}$%
%
%
%
\thanks{
$^{*}$Ye Yuan, Kehan Chen, and Xinqiang Yu contributed
equally to this work.
$^{\dagger}$Jiawei He
({\tt\footnotesize jwhe2024@gmail.com}) and Zhulin An
({\tt\footnotesize anzhulin@ict.ac.cn}) are the corresponding authors.
}%
\thanks{
$^{1}$Ye Yuan and Wentao Xu are with the School of
Information Science and Technology, ShanghaiTech University,
Shanghai 201210, China, and also with the Institute of
Computing Technology, Chinese Academy of Sciences,
Beijing 100190, China
({\tt\footnotesize
\{yuanye2024,xuwt2024\}@shanghaitech.edu.cn}).
}%
\thanks{
$^{2}$Libo Huang, Chuanguang Yang, and Zhulin An are with
the State Key Laboratory of AI Safety, Institute of Computing
Technology, Chinese Academy of Sciences,
Beijing 100190, China({\tt\footnotesize
\{yangchuanguang,anzhulin\}@ict.ac.cn};
{\tt\footnotesize www.huanglibo@gmail.com}).
}%
\thanks{
$^{3}$Kehan Chen, Xinqiang Yu, and Yan Huang are with
the National Laboratory of Pattern Recognition,
Institute of Automation, Chinese Academy of Sciences,
Beijing, China. Kehan Chen and Xinqiang Yu are also with
the School of Artificial Intelligence, University of Chinese
Academy of Sciences, Beijing, China
({\tt\footnotesize
\{xinqiang.yu,yhuang\}@nlpr.ia.ac.cn};
{\tt\footnotesize kehan.chen@cripac.ia.ac.cn}).
}%
\thanks{
$^{4}$Ye Yuan, Heng Wang, and Jiawei He are with
XYZ Embodied AI, Beijing, China.
}%
\thanks{
Digital Object Identifier (DOI): 10.1109/LRA.2026.3746066.
}%
}


\maketitle

\begin{abstract}
City-scale outdoor navigation is currently hindered by the heavy reliance on dense maps or costly navigation supervision. In this work, we introduce a novel paradigm for leveraging directional instructions from commercial navigation tools (e.g., Google Maps). To bridge the gap between commercial instructions and executable navigation actions, while mitigating long-horizon error accumulation through robust trajectory recovery, we propose DA-Nav, a Direction-Aware vision-language Navigation framework that reformulates navigation as a discrete spatial grounding problem on the egocentric 2D image plane. To achieve trajectory recovery, DA-Nav employs a Chain-of-Thought (CoT) reasoning process encompassing deviation assessment, action prediction, and target grid selection. We further introduce ReDA, a dataset that provides direction-aware instructions and recovery trajectories to enhance spatial grounding and support CoT recovery reasoning. Extensive experiments in CARLA demonstrate that DA-Nav achieves a high success rate of \textbf{56.16\%} in unseen urban environments, outperforming existing State-of-The-Art (SoTA) methods while maintaining a substantially stronger recovery capability. Furthermore, without fine-tuning, DA-Nav seamlessly adapts to both quadruped and humanoid robots, enabling stable kilometer-scale closed-loop outdoor navigation in complex real-world environments.
\end{abstract}

\begin{IEEEkeywords}
Vision-Based Navigation, Data Sets for Robot Learning, Spatial Grounding, Trajectory Recovery, Sim-to-Real Transfer
\end{IEEEkeywords}

\begin{figure*}[t] 
    \centering
    \includegraphics[width=0.9\textwidth]{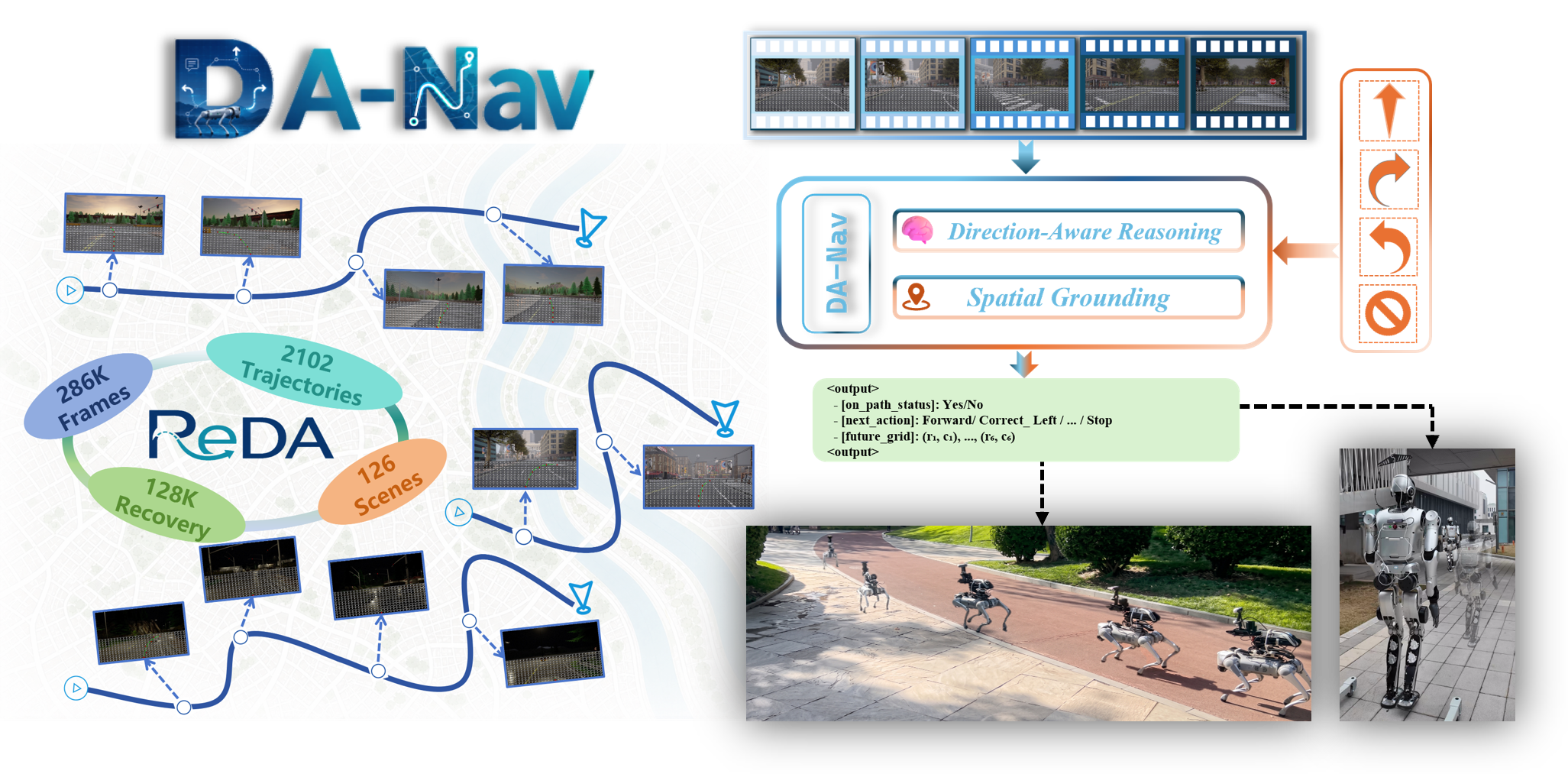} 
    \vspace{-0.7cm}
    \caption{Overview of DA-Nav. Trained on the ReDA dataset, our policy reformulates outdoor navigation as a vision-language-conditioned discrete spatial grounding problem on the egocentric image plane. The model autoregressively predicts structured decisions and image-plane trajectories, achieving robust zero-shot sim-to-real deployment across diverse legged robots.}
    \label{fig:teaser}
    \vspace{-0.6cm}
\end{figure*}

\section{Introduction}
\IEEEPARstart{C}{ity-scale} outdoor mobile robots hold significant potential for applications such as logistics delivery, automated inspection, and public services. However, reliable long-horizon urban navigation remains challenging. Existing approaches typically rely on  Simultaneous Localization and Mapping (SLAM) and global path planning~\cite{SLAM_Review, LidarSLAM_2024}. In city-scale scenarios, however, such methods often incur substantial mapping and maintenance costs and remain sensitive to dynamic environmental changes. To reduce the reliance on dense maps, recent studies have explored navigation using coarse-grained guidance signals such as natural language instructions, GPS waypoints, and semantic landmarks~\cite{LMNav, Touchdown, ViNT, NaVid, NoMaD, CityWalker}. Nevertheless, these approaches usually require dense navigation supervision, such as fine-grained language instruction and landmark annotations\cite{cui2026fcanig}, or manually collected trajectories, which limits scalability in city-scale deployments.

To address these limitations, we investigate a new paradigm for city-scale navigation by leveraging instructions from commercial navigation tools (e.g., Google Maps, Amap), formulating it as a Direction-Aware Vision-Language Navigation (VLN) task. Compared with manually constructed dense navigation supervision, commercial navigation tools already provide mature global path planning and real-time local directional guidance. However, such instructions are designed for human cognition and only contain coarse-grained directional information (e.g., “turn right in 50 meters”), making them unsuitable for direct robotic execution. Therefore, the first challenge in this paradigm is how to ground sparse directional instructions with egocentric observations into locally executable motion behaviors. Recent development of Vision-Language Models (VLM)~\cite{SpatialVLM, palme} and Vision-Language-Action (VLA) models~\cite{OpenVLA2024, RT2} demonstrate the potential for solving the cross-modal alignment problem. Moreover, during long-horizon navigation, perception and control errors inevitably accumulate over time, causing the robot to gradually deviate from the intended path. Consequently, robust recovery from trajectory deviations is another critical challenge.

To this end, we propose a \textbf{D}irection-\textbf{A}ware \textbf{Nav}igation (\textbf{DA-Nav}) method based on VLM for city-scale navigation. Unlike conventional continuous trajectory regression methods~\cite{diffusionpolicy, NoMaD, CityWalker}, DA-Nav reformulates navigation as a vision-language-conditioned discrete spatial grounding problem on the 2D image plane. As illustrated in Fig.~\ref{fig:teaser}, we construct a discrete spatial grid over egocentric visual observations and leverage the spatial grounding capability of VLMs to predict local navigation targets directly on the image plane. Furthermore, DA-Nav introduces a structured sequential reasoning process that decomposes navigation into a series of decision steps, including deviation assessment, action prediction, and image-plane target selection. Specifically, the model first determines whether trajectory deviation has occurred, then predicts the corresponding navigation or corrective action, and finally generates the target grid sequence on the image plane. Implemented in a Chain-of-Thought (CoT) formulation~\cite{mapgpt, navgpt, cotvlnbench, tcot, clcotnav}, this structured decision process enables trajectory recovery during long-horizon navigation.

In addition, enabling direction-aware navigation and trajectory recovery requires training data with directional instructions and recovery behaviors. However, the instructions in existing outdoor navigation datasets are predominantly landmark-driven rather than direction-aware instructions~\cite{Touchdown, LMNav, Talk2Nav}. Besides, current datasets are limited to expert trajectories and lack supervision for recovery behaviors~\cite{Anderson2018VLN, Krantz2020VLNCE, CityWalker}. Therefore, we introduce the \textbf{ReDA} dataset, including both \textbf{Re}covery trajectories and \textbf{D}irection-\textbf{A}ware instructions. Specifically, we develop an automated data pipeline to generate directional instructions across 2,102 trajectories within 126 scenes in the CARLA simulator~\cite{CARLA}. Besides, we actively inject perturbations to generate deviated trajectories and construct corresponding CoT reasoning steps for training recovery capability.

To validate the effectiveness of the proposed approach, we conducted extensive closed-loop evaluations in the CARLA simulator. DA-Nav achieved a navigation success rate of 56.16\% in unseen urban environments, outperforming other State-of-The-Art (SoTA) methods. Furthermore, the learned policy achieves zero-shot transfer to real-world robots, including a Unitree Go2 quadruped robot and a Leju Kuavo-V humanoid robot, without any real-world fine-tuning, enabling stable kilometer-scale closed-loop outdoor navigation.
Our main contributions are summarized as follows:
\begin{itemize}
   \item We propose DA-Nav, a VLM-based long-horizon outdoor navigation model driven by direction-aware instructions.  It performs discrete image-plane spatial grounding and CoT reasoning for trajectory recovery.
   \item We introduce ReDA, a dataset comprising direction-aware instructions and trajectory recovery annotations, enabling robust learning under Out-Of-Distribution (OOD) deviation states.
   \item DA-Nav achieves SoTA performance in CARLA. Notably, it demonstrates exceptional zero-shot sim-to-real transfer, achieving stable long-horizon navigation over 1.2 km in real-world environments.
\end{itemize}

\section{Related Work}
\label{sec:related_work}

\subsection{Outdoor Navigation in Unstructured Environments}
To bypass the prohibitive mapping costs and dynamic sensitivity inherent in conventional SLAM-based systems~\cite{SLAM_Review, LidarSLAM_2024}, recent approaches have explored high-level guidance such as GPS waypoints, topological graphs, and semantic landmarks~\cite{ViNT, NoMaD, CityWalker}. However, maintaining reliable localization and semantic consistency in large unstructured outdoor environments remains challenging due to GPS multi-path errors and dynamic scene variations. While conventional VLN methods rely on fine-grained instructions to describe specific paths~\cite{LMNav}, such dense supervision is unrealistic for city-scale deployment. In contrast, commercial navigation tools provide lightweight coarse-grained guidance for long-horizon navigation, but such instructions lack the precision required for direct robotic control. DA-Nav addresses this challenge by grounding sparse directional guidance into reliable local spatial decisions.

\vspace{-0.3cm}
\subsection{Vision-Language Models for Embodied Navigation}
The strong semantic understanding and cross-modal alignment capabilities of VLMs have driven a paradigm shift in embodied navigation~\cite{song2023llm, palme, navfom, omnivla}. Beyond model capacity, VLN generalization is also affected by dataset bias and spurious correlations~\cite{CausalVLN}, while sim-to-real transfer introduces additional visual and physical disparities~\cite{EmbodiedGap}. While end-to-end VLA models such as RT-2~\cite{RT2} and OpenVLA~\cite{OpenVLA2024} exhibit impressive generalization, modeling continuous spatial actions in an autoregressive manner remains challenging for stable long-horizon execution~\cite{SpatialVLM, RTTrajectory}. Recent studies suggest that VLMs are more effective at reasoning over 2D visual representations than regressing 3D waypoints~\cite{SpatialVLM, navitrace, SpatialVLA}. This indicates that VLM-based navigation instability largely stems from a discrepancy between model capability and action representation. Motivated by this, DA-Nav reformulates navigation planning as a vision-language-conditioned discrete spatial grounding problem. By projecting candidate targets onto a discretized egocentric image grid and performing decision-making directly on the image plane, DA-Nav aligns the action space with the visual reasoning characteristics of VLMs, thereby mitigating spatial hallucination and improving navigation stability.

\vspace{-0.20cm}
\subsection{Closed-Loop Error Recovery and Reasoning}
Most existing vision-based navigation policies are trained via behavior cloning on nominal expert-only demonstrations~\cite{ViNT, diffusionpolicy}. While effective in offline settings, this paradigm inevitably suffers from covariate shift during closed-loop execution~\cite{ross2011dagger, codevilla2019exploring, care2025}. Minor execution errors can quickly push the agent into OOD states, leading to compounding deviations and eventual failure. Interactive approaches such as DAgger~\cite{ross2011dagger} mitigate this issue by querying expert supervision online, but they require expensive simulator-in-the-loop setups, making them difficult to scale to large VLM-based systems. Recent works have explored structured reasoning and decision decomposition to improve interpretability in embodied systems~\cite{ahn2022can, wake2023chatgpt}. However, most existing approaches focus primarily on high-level task planning, with limited consideration of low-level execution assessment and recovery under deviation states. In contrast to prior approaches that primarily rely on expert-only demonstrations or high-level planning supervision, DA-Nav incorporates corrective supervision and direction-aware CoT reasoning for robust long-horizon closed-loop execution.

\section{Task Definition and ReDA Dataset}

\subsection{Direction-Aware VLN }

In our task, at each time step $t$, the robot receives a sequence of egocentric RGB observations $\mathbf{O}_t = (o_{t-k}, \dots, o_t)$ with $k=4$, together with a discrete directional instruction $I_t \in \{\texttt{FORWARD}, \texttt{TURN\_LEFT}, \texttt{TURN\_RIGHT}, \texttt{STOP}\}$. In simulation, $I_t$ is generated from the global route produced by CARLA's route planner, whereas during real-world deployment, it is parsed from instructions provided by commercial navigation tools. Given the multimodal context $C_t = \{\mathbf{O}_t, I_t\}$, the goal is to learn a policy $\pi_\theta$ that predicts a feasible local navigation trajectory. Unlike conventional approaches that directly regress continuous 3D waypoints, we reformulate navigation as a vision-language-conditioned discrete spatial grounding problem on the egocentric image plane. Future trajectories are represented as discretized candidate locations on the image plane.

\vspace{-0.3cm}
\subsection{Dataset Description}
\label{subsec:dataset_description}

\begin{table}[t]
\vspace*{0.3cm}
\caption{Comparison of ReDA with Representative Vision-Language Navigation Datasets}
\vspace{-0.3cm}
\label{tab:dataset_comparison}
\centering
\resizebox{0.98\columnwidth}{!}{%
\begin{tabular}{l c c l l}
\toprule
\textbf{Dataset} & \textbf{Environment} & \textbf{Depth} & \textbf{Instruction Type} & \textbf{Action Space} \\
\midrule
R2R~\cite{Anderson2018VLN}$^\ast$      & Indoor  & $\times$   & Fine-grained Description
     & Graph Nodes    \\
Touchdown~\cite{Touchdown}$^\ast$      & Outdoor & $\times$   & Fine-grained Description     & Graph Nodes    \\
VLN-CE~\cite{Krantz2020VLNCE}$^\ast$   & Indoor  & \checkmark & Fine-grained Description     & 3D Control     \\
Talk2Nav~\cite{Talk2Nav}$^\ast$        & Outdoor & $\times$   & Fine-grained Description     & 3D Waypoints   \\
LM-Nav~\cite{LMNav}$^\ast$             & Outdoor & $\times$   & Landmark-based     & Graph Nodes    \\
GNM~\cite{Shah2023GNM}$^\ast$          & Mixed   & $\times$   & Image Goal         & 2D Waypoints   \\
CityWalker~\cite{CityWalker}$^\ast$    & Outdoor & $\times$   & Path-level        & 3D Waypoints   \\
NaVid~\cite{NaVid}$^\ast$              & Indoor  & $\times$   & Path-level        & Discrete Commands \\
NaVILA~\cite{NaVILA}$^\ast$            & Mixed   & \checkmark & Mixed             & Discrete Commands \\
\midrule
\textbf{ReDA (Ours)}$^\dagger$         & \textbf{Outdoor} & \textbf{$\times$} & \textbf{Direction-based} & \textbf{2D Image Grid} \\
\bottomrule
\multicolumn{5}{p{\columnwidth}}{\footnotesize
\textit{Note}: All datasets use RGB input.
$^\ast$ expert-only; $^\dagger$ expert + recovery data.}
\end{tabular}%
}
\vspace{-0.4cm}
\end{table}

As summarized in Table~\ref{tab:dataset_comparison}, the representative datasets cover diverse instruction and action representations, but do not jointly provide direction-based guidance, trajectory labels on an image grid, and recovery demonstrations. To address this gap, we construct ReDA to support both spatial grounding of coarse directional instructions and corrective behavior learning during closed-loop navigation. ReDA contains approximately 286k sequential samples generated from 2,102 trajectories in CARLA Town01--05 and Town10HD, including 158k expert navigation frames and 128k recovery frames. The 239 evaluation routes were constructed separately from the ReDA trajectories and form the benchmark described in Sec.~\ref{subsec:simulation}. The two core components of ReDA, namely the \textit{Egocentric Grid Trajectory Representation} and the \textit{Direction-Aware Recovery Data Generation}, are detailed in the following subsections.

\vspace{-0.3cm}
\subsection{Egocentric Grid Trajectory Representation}
To instantiate this direction-aware spatial selection problem, we construct a discrete grid on the egocentric RGB image plane and define the valid spatial region as:
\begin{equation}
    G = \{ (r, c) \mid r \in [13, 23], c \in [0, 28] \} \subset \mathbb{Z}^2,
\end{equation}
where $r$ and $c$ denote the row and column indices, corresponding to forward and lateral directions, respectively. The valid region excludes sky and distant background areas, focusing on traversable regions in front of the robot. Based on this representation, we discretize the future trajectory into a sequence of grid locations with a horizon length of $L = 6$:
\begin{equation}
    \mathbf{P}_t = (\mathbf{g}_{t+1}, \dots, \mathbf{g}_{t+L}), \quad \mathbf{P}_t \in \mathbb{R}^{L \times 2}.
\end{equation}
This sequence represents the robot motion over the next 3 seconds sampled at 2 Hz. This representation constrains trajectory prediction to discrete image-space locations.

Building upon this grid representation, we further construct a structured CoT decision sequence $Y_t = (s_t, c_t, \mathbf{P}_t)$. Here, $s_t \in \{\texttt{Yes}, \texttt{No}\}$ serves to explicitly evaluate whether the robot has deviated from the reference path; $c_t \in \{$ \texttt{FORWARD}, \allowbreak \texttt{TURN\_LEFT}, \allowbreak \texttt{TURN\_RIGHT}, \allowbreak \texttt{CORRECT\_LEFT}, \allowbreak \texttt{CORRECT\_RIGHT}, \allowbreak \texttt{STOP} $\}$ serves as the CoT action; and $\mathbf{P}_t$ denotes the predicted spatial grid sequence.

\vspace{-0.3cm}
\subsection{Direction-Aware Recovery Data Generation}

To expand the state distribution, we propose an automated pipeline in the CARLA simulator governed by a three-state Finite State Machine (FSM): \textsc{Stable}, \textsc{Drifting}, and \textsc{Recovering}. In the \textsc{Stable} state, the agent tracks the expert trajectory. To induce OOD states, steering perturbations trigger the \textsc{Drifting} state. Once the lateral error exceeds a safety threshold ($e_y \ge 0.35$~m), the FSM switches to the \textsc{Recovering} state. During recovery, the agent adopts an adaptive tracking strategy in which the look-ahead distance $l_d$ is proportional to $e_y$, enabling smooth trajectory correction. To avoid learning unstable behaviors, frames from the \textsc{Drifting} state are discarded.

For the retained frames, we synthesize discrete navigation instructions using a deterministic rule based on the agent's turning state and motion status along the CARLA reference route, including \textsc{Forward}, \textsc{Turn\_Left}, \textsc{Turn\_Right}, and \textsc{Stop}. The ground-truth deviation label $s_t^*$ is derived from the active FSM state, while the corrective action $c_t^*$ is determined by the sign of the lateral deviation $e_y$ during recovery. Finally, future expert 3D waypoints are smoothed via cubic spline interpolation, projected onto the egocentric image plane using the calibrated camera projection model, and discretized into trajectory labels $\mathbf{P}_t^*$.

\begin{figure}[t] 
\vspace*{0.3cm}
    \centering
    \includegraphics[width=0.96\columnwidth]{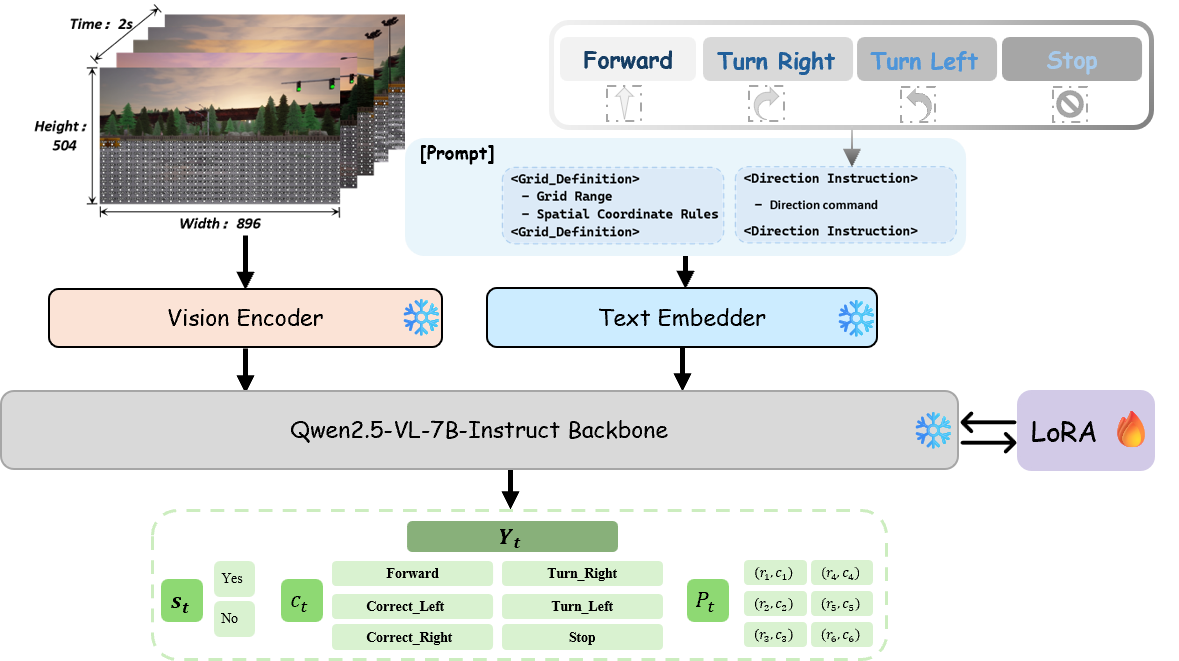}
    \caption{Overview of the DA-Nav architecture. Multimodal inputs are processed by frozen encoders and passed to a LoRA-finetuned Qwen2.5-VL-7B backbone. The model autoregressively outputs a structured CoT decision sequence consisting of state assessment, corrective action, and discrete future trajectories on the image grid.}
    \label{fig:system_architecture}
    \vspace{-0.5cm}
\end{figure}

\section{Methodology}
As shown in Fig.~\ref{fig:system_architecture}, DA-Nav takes temporal egocentric observations and sparse directional instructions as input and reformulates long-horizon navigation as a vision-language-conditioned discrete spatial grounding problem. Specifically, a structured VLM policy autoregressively predicts a CoT decision sequence together with discrete image-plane navigation targets, which are subsequently projected into the robot body frame and executed through a furthest-point control interface for robust outdoor navigation.

\subsection{DA-Nav}
We instantiate the policy $\pi_\theta$ using the pre-trained Qwen2.5-VL-7B-Instruct backbone. At each timestep, the model takes a multimodal context $C_t$ as input. The visual history $\mathbf{O}_t$ is encoded by a frozen visual encoder and projected into the language embedding space via a lightweight adapter, allowing seamless integration with the textual instruction $I_t$. 

To improve spatial reasoning under trajectory deviations, we adopt a structured prompting strategy with output constraints.  First, we inject spatial grounding constraints by encoding the discretized image grid $G$ in the prompt, establishing the correspondence between grid indices and spatial locations. Second, we enforce logical decomposition by structuring the output generation as a sequential decision process, where the model first performs state assessment, then selects an action, and finally predicts the trajectory. This formulation encourages the VLM to explicitly evaluate the navigation state $s_t$ and infer a high-level action $c_t$ before predicting the spatial trajectory $\mathbf{P}_t$, improving the consistency between CoT decisions and spatial trajectory prediction.

The policy is trained as an autoregressive generation process over the decision sequence $Y_t$. Given the context $C_t$, the policy factorizes as:
\begin{equation}
\pi_\theta (Y_t \mid C_t) = \prod_{j=1}^{|Y_t|} P_\theta \left(y_t^{(j)} \mid y_t^{(<j)}, C_t \right),
\end{equation}
$\pi_\theta$ is optimized using a standard next-token prediction objective with supervision from the ReDA dataset. 

To adapt the pre-trained VLM without catastrophic forgetting, we apply LoRA to the attention blocks, keeping the visual encoder and base LLM weights frozen. This parameter-efficient strategy enables the model to internalize corrective patterns from the ReDA dataset while preserving its inherent cross-modal generalization capabilities.

\subsection{Image-Plane-to-Body-Frame Spatial Grounding}

\begin{figure}[htbp]
    \centering

    \includegraphics[width=0.9\columnwidth]{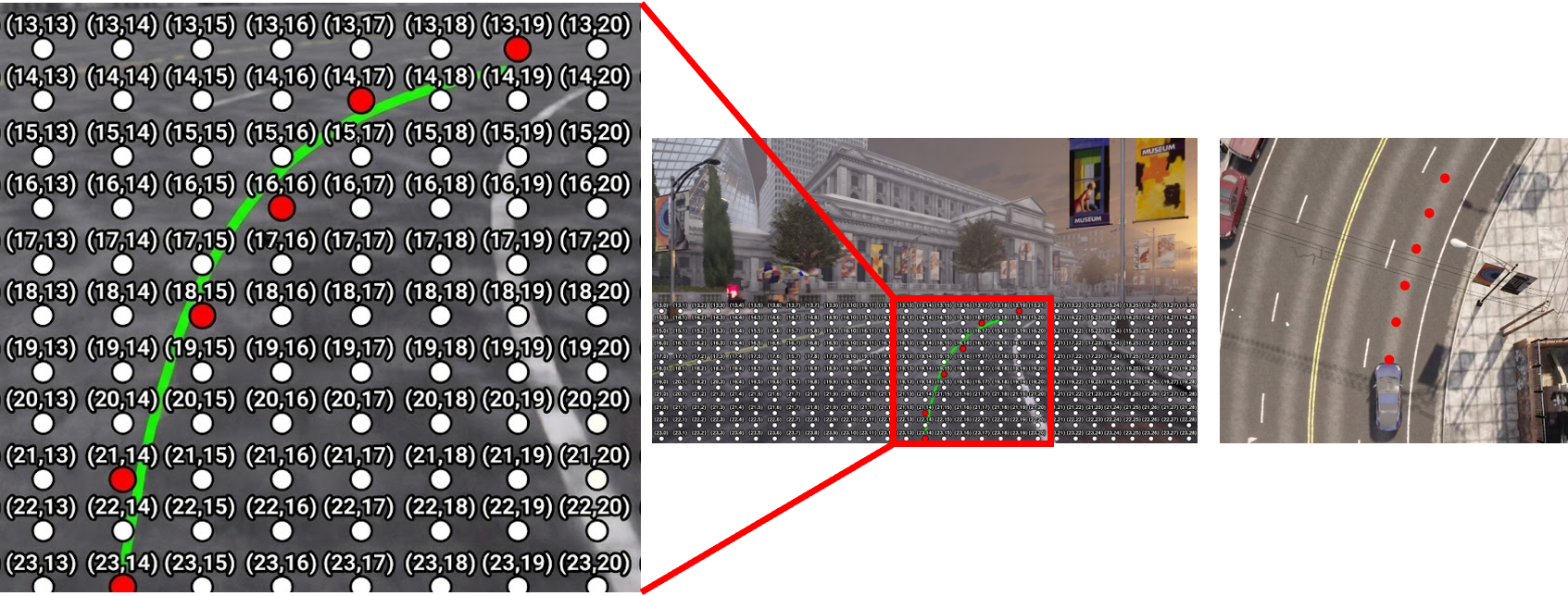}
    \caption{Visualization of the egocentric grid representation and spatial grounding process. \textbf{Left and Middle}: The traversable region in the RGB observation is discretized into a 2D image-plane grid, where future trajectories are represented as discrete targets $\mathbf{P}_t$ (red dots). \textbf{Right}: The predicted targets are projected into continuous local-body-frame trajectories $\mathbf{W}_t$ for robot execution.}
    \label{fig:spatial_projection}
\end{figure}

To bridge the gap between discrete image-space predictions and continuous physical execution, the geometric control interface first transforms the predicted pixel grid sequence $\mathbf{P}_t$ into a 3D continuous trajectory $\mathbf{W}_t = (\mathbf{w}_{t+1}, \dots, \mathbf{w}_{t+L})$ in the robot's local body frame, where $w_k = \left( x_{\text{body}}^{(k)}, y_{\text{body}}^{(k)} \right)$, as illustrated in Fig.~\ref{fig:spatial_projection}.

To ensure robustness across diverse outdoor environments, we employ a hybrid spatial projection strategy. When reliable depth observations are available, the predicted image-plane targets are projected into metric coordinates using the camera intrinsic matrix and local depth statistics. If depth sensing becomes unreliable, the system falls back to an Inverse Perspective Mapping (IPM) strategy under a flat-ground assumption. This hybrid design ensures continuous and reliable spatial grounding across varying sensing conditions.

\subsection{Furthest-Point Control Interface}
Due to the latency of autoregressive VLM inference, directly tracking dense predicted trajectories may introduce high-frequency control fluctuations during closed-loop execution. To overcome this, we adopt a furthest-point target pursuit strategy. Instead of tracking all predicted waypoints, we select the furthest waypoint in the prediction horizon as the control target, i.e., $\mathbf{w}_{\text{target}} = \mathbf{w}_{t+L}$. Using the furthest predicted waypoint as the control target provides a temporally smoother furthest-point signal and reduces sensitivity to short-term prediction noise.

Guided by this furthest-point target, an adaptive low-level controller computes the final motion commands. A heading-error-driven controller computes the angular velocity:
\begin{equation}
\omega_z = \mathrm{sat}_{\omega_{\max}} \left( k_{\text{steer}} \cdot \arctan2(y_{\text{target}}, x_{\text{target}}) \right),
\end{equation}
where $\mathrm{sat}_{\omega_{\max}}(\cdot)$ denotes saturation to enforce actuator limits. Furthermore, the linear velocity is adaptively adjusted based on the steering magnitude:
\begin{equation}
v_{\text{target}} =
\begin{cases}
v_{\text{nom}}, & |\omega_z| \leq \omega_{\text{th}} \\
v_{\text{low}}, & |\omega_z| > \omega_{\text{th}}.
\end{cases}
\end{equation}
This formulation automatically restricts aggressive steering via limits and decelerates the robot during sharp maneuvers, improving kinematic stability and enabling smooth platform adaptation during closed-loop deployment. \revision{Since the task targets forward navigation on urban roads and pedestrian paths, DA-Nav handles sharp turns through continuous curved motion and does not include an explicit action for pure in-place rotation.}

\section{Experiments}

\subsection{Experimental Setup}
\label{subsec:setup}

To comprehensively evaluate our method, we compare DA-Nav against five representative baselines. For end-to-end foundation models, we select \textit{CityWalker}~\cite{CityWalker}, a Transformer-based waypoint regression policy, and \textit{ViNT}~\cite{ViNT}, a goal-conditioned topological navigation model. To assess large vision-language capabilities, we include \textit{NaVid}~\cite{NaVid}, which directly deduces physical actions from video streams, alongside the unmodified \textit{Zero-shot Qwen2.5-VL-7B}~\cite{qwen2025vl} to isolate the impact of our task-specific fine-tuning. Finally, we compare against \textit{NaVILA}~\cite{NaVILA}, a hierarchical VLA system that employs a VLM for mid-level navigation instruction generation coupled with low-level execution.

To compare the baselines using the same training data, CityWalker, ViNT, NaVid, and NaVILA are fine-tuned on the ReDA training split before closed-loop evaluation. All runs start from official pretrained weights and retain each baseline's architecture, input processing, action representation, and training objective. We adapt only the data interfaces required by each original training paradigm. ViNT receives moving-horizon local goal images with 5~m lookahead, while CityWalker receives deterministic sparse waypoints. NaVid and NaVILA use identical egocentric visual histories and unified low-level proportional controllers. Goal termination relies on visual recognition of semantic destinations (e.g., a distinct red vehicle). Zero-shot Qwen2.5-VL uses its official weights with the same system settings.

\begin{figure*}[t]
\vspace*{0.3cm}
  \centering
  \includegraphics[width=0.85\linewidth, keepaspectratio]{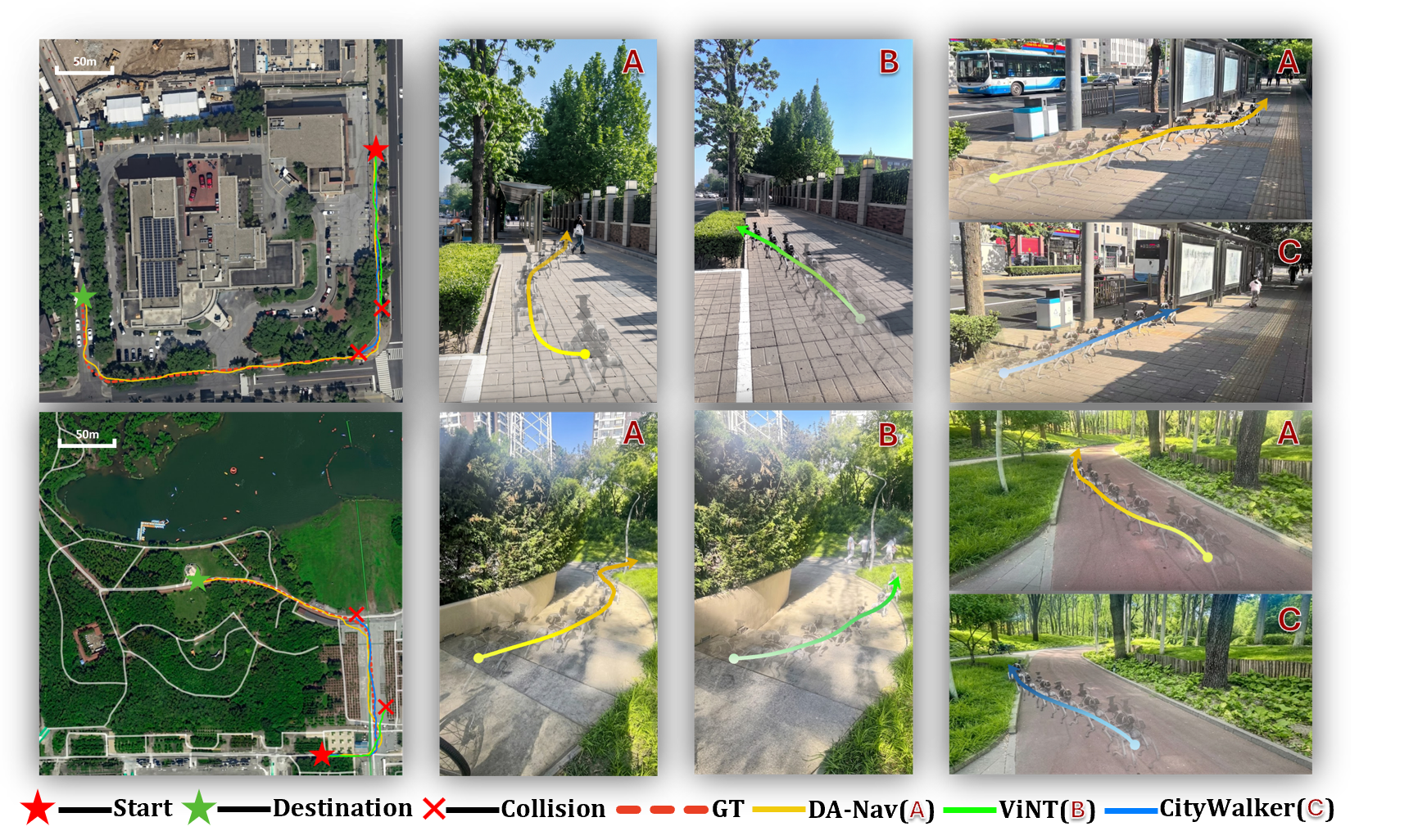}
  \vspace{-0.60cm}
  \caption{Real-world closed-loop navigation comparison. Satellite maps denote the start (red star), destination (green star), and reference path (red dashed line). Third-person views visualize the execution trajectories of DA-Nav (A, yellow), ViNT (B, green), and CityWalker (C, blue). Red crosses ($\times$) indicate collisions during execution.}
\label{fig:real_world_deploy}
\vspace{-0.60cm}
\end{figure*}

For evaluation, in addition to standard embodied navigation metrics including Success Rate (SR), Route Completion (RC), and Success weighted by Path Length (SPL), we note that these conventional metrics fail to adequately capture the model's recovery capability under trajectory deviations. To more precisely evaluate autonomous correction, we additionally introduce two recovery-oriented metrics: Deviation Frequency (DF) and Correction Success Rate (CSR). A deviation event is counted once when the lateral error first exceeds the safety threshold after returning to a nominal state. Let $E$ denote the set of such deviation events where the lateral error $e_y$ exceeds a safety threshold $\delta_{\text{safe}}$ (e.g., $0.35$\,m), $D_{\text{total}}$ the total traveled distance (in meters), and $E_{\text{success}} \subseteq E$ the subset of events successfully reconverging to $e_y < \delta_{\text{safe}}$ without collision. These metrics are formally defined as:
\begin{equation}
    \text{DF} = \frac{|E|}{D_{\text{total}}} \times 100, \qquad \text{CSR} = \frac{|E_{\text{success}}|}{|E|} \times 100\%.
\end{equation}

\subsection{Quantitative Benchmark in Simulation}
\label{subsec:simulation}

We additionally construct a custom evaluation benchmark of 239 long-horizon navigation routes, separate from the CARLA Leaderboard. It includes 93 routes on maps seen during ReDA construction (Town03, Town04, Town05, and Town10HD) and 146 routes on unseen maps (Town06, Town07, and Town15). The evaluation routes were constructed independently of the ReDA trajectories. No evaluation route duplicates a complete ReDA trajectory or uses the same start-goal pair as a ReDA trajectory. Dynamic traffic, pedestrians, and varied weather and lighting conditions are used during both data collection and evaluation.

\begin{figure}[t]
\vspace*{0.3cm}
    \centering
   
    \includegraphics[width=0.8\columnwidth]{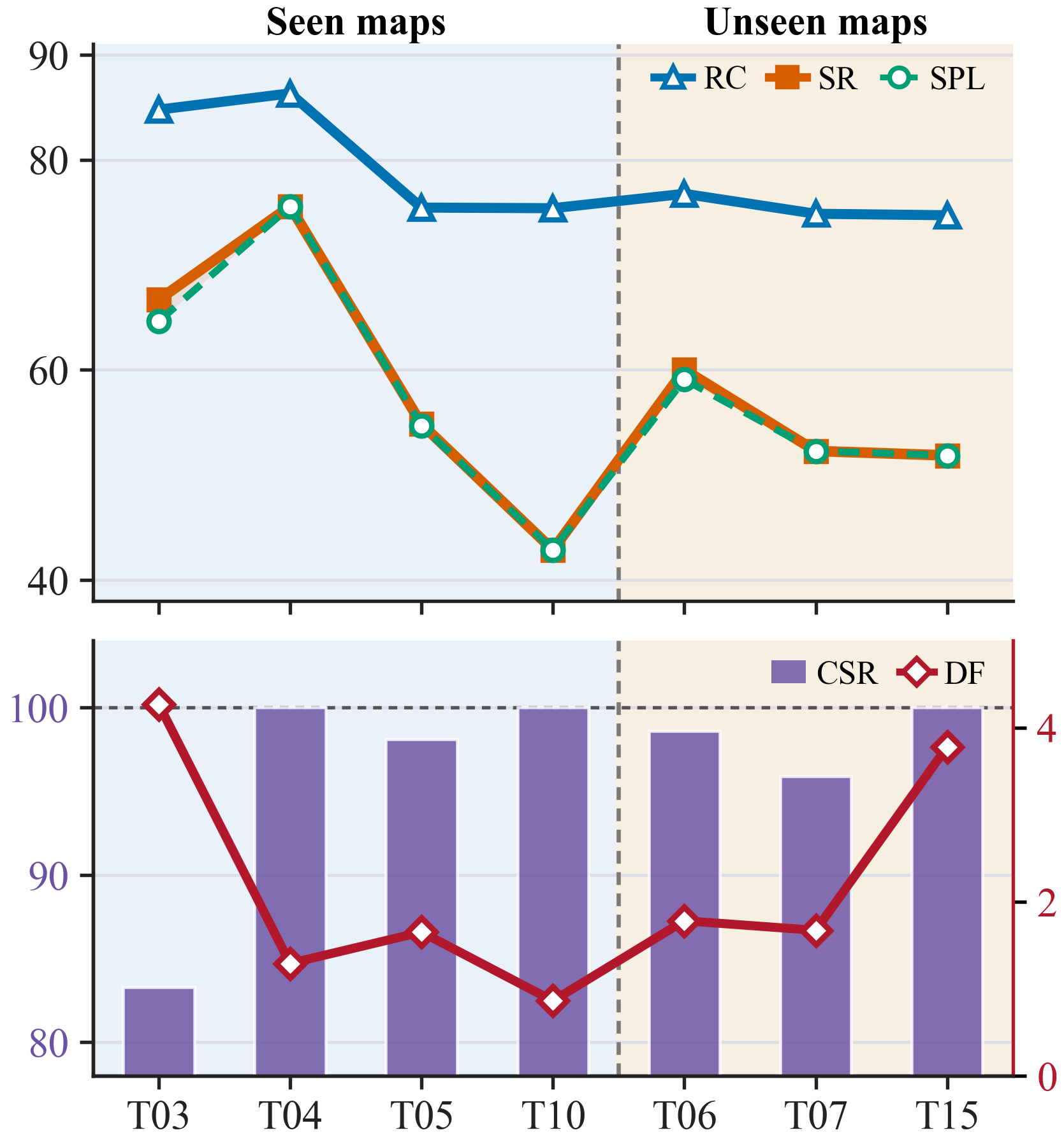}
    \vspace{-0.2cm}
    \caption{Per-town evaluation of DA-Nav across seen and unseen environments. \textbf{Top:} Standard navigation metrics (RC, SR, SPL) across different CARLA towns. \textbf{Bottom:} Relationship between deviation frequency (DF, red line) and correction success rate (CSR, purple bars).}
    \label{fig:per_town_analysis}
    \vspace{0.1cm}
\end{figure}

All methods are evaluated on this benchmark. Quantitative results are reported in Table~\ref{tab:global_benchmark}, with per-town generalization results shown in Fig.~\ref{fig:per_town_analysis}. DA-Nav achieves the best overall performance, obtaining an SR of 59.00\%, SPL of 58.66, and CSR of 98.15\%. Moreover, its SR decreases by only 7.28 percentage points when transferred from seen to unseen environments, demonstrating robust spatial generalization and trajectory recovery under distribution shifts.

\begin{table}[t]
\vspace*{0.3cm}
\caption{Quantitative Comparison on 239 Evaluation Routes. The dagger denotes ReDA-fine-tuned baselines}
\vspace{-0.3cm}
\label{tab:global_benchmark}
\centering
\resizebox{\columnwidth}{!}{%
\begin{tabular}{c ccccc}
\toprule
\textbf{Method} & \textbf{SR (\%)} $\uparrow$ & \textbf{RC (\%)} $\uparrow$ & \textbf{SPL} $\uparrow$ & \textbf{DF} $\downarrow$ & \textbf{CSR (\%)} $\uparrow$ \\
\midrule
CityWalker$^\dagger$~\cite{CityWalker}  & 41.42 & 73.54 & 41.33 & 2.19 & 36.93 \\
ViNT$^\dagger$~\cite{ViNT}        & 52.72 & 75.38 & 52.00 & 1.72 & 44.65 \\
NaVid$^\dagger$~\cite{NaVid}       & 17.99 & 64.85 & 17.99 & 0.92 & 19.71  \\
NaVILA$^\dagger$~\cite{NaVILA}    & 22.18 & 78.61 & 22.18 & 0.42 & 15.28  \\
Zero-shot Qwen2.5-VL~\cite{qwen2025vl}  & 11.30 & 43.21 & 11.30 & 3.42 & 42.76 \\
\midrule
\textbf{DA-Nav (Ours)} & \textbf{59.00} & 77.82 & \textbf{58.66} & 1.85 & \textbf{98.15} \\
\bottomrule
\end{tabular}%
}
\vspace{0.5mm}
\begin{minipage}{\columnwidth}
\footnotesize
$^\dagger$ Fine-tuned on the same ReDA training split while
retaining the baseline's original architecture, action representation,
and training objective.
\end{minipage}
\vspace{-0.4cm}
\end{table}

Table~\ref{tab:global_benchmark} reports the ReDA-fine-tuned results for CityWalker, ViNT, NaVid, and NaVILA, while Zero-shot Qwen2.5-VL is evaluated using its official weights without ReDA fine-tuning. The fine-tuned baselines show distinct failures in long-horizon execution. CityWalker and ViNT obtain relatively high SR and SPL but limited CSR of 36.93\% and 44.65\%, indicating weak recovery. NaVid and NaVILA achieve lower SR and SPL, with CSR values of only 19.71\% and 15.28\% despite low DF values of 0.92 and 0.42. Thus, fewer deviations do not necessarily imply successful recovery. NaVILA's RC of 78.61\% further shows that route completion alone does not guarantee successful navigation. Even after ReDA fine-tuning, the baselines struggle to combine stable grounding, reliable navigation, and recovery. With structured reasoning, image-space grounding, and recovery, DA-Nav achieves an SR of 59.00\%, an SPL of 58.66, and a CSR of 98.15\% under the same training-data condition.

\subsection{Ablation Studies}
\label{subsec:ablation}

To examine the contributions of structured reasoning and recovery supervision, we compare DA-Nav with two variants on the same 239 routes. One directly predicts spatial targets without CoT reasoning, while the other is trained only on expert trajectories without recovery data. We also retrain DA-Nav with coarse $16\times17$ and fine $32\times35$ grids using regenerated trajectory labels. We further evaluate robustness to directional guidance by corrupting 5\% of the instructions (\textit{Noisy}) or increasing their update interval from 0.25\,s to 1\,s (\textit{Coarse}). All other settings remain unchanged.

\begin{table}[t]
\caption{Ablation and Robustness on 239 Evaluation Routes}
\vspace{-0.3cm}
\label{tab:ablation}
\centering
\resizebox{\columnwidth}{!}{
\begin{tabular}{c ccccc}
\toprule
\textbf{Model Variant} & \textbf{SR (\%)} $\uparrow$ & \textbf{RC (\%)} $\uparrow$ & \textbf{SPL} $\uparrow$ & \textbf{DF} $\downarrow$ & \textbf{CSR (\%)} $\uparrow$ \\
\midrule
w/o Recovery Data & 29.71 & 62.01 & 29.68 & \textbf{1.31} & 15.46 \\
w/o CoT Reasoning & 38.91 & 68.37 & 38.91 & 4.30 & 50.11 \\
Fine Grid & 26.78 & 65.48 & 26.75 & 2.88 & 83.28 \\
Coarse Grid & 48.12 & 73.86 & 48.10 & 1.91 & 70.44 \\
Noisy Direction & 51.50 & 74.45 & 51.12 & 2.35 & 97.62 \\
Coarse Direction & 50.20 & 73.93 & 49.95 & 1.86 & 97.91 \\
\midrule
\textbf{DA-Nav (Full)} & \textbf{59.00} & \textbf{77.82} & \textbf{58.66} & 1.85 & \textbf{98.15} \\
\bottomrule
\end{tabular}
}
\vspace{-0.4cm}
\end{table}

As shown in Table~\ref{tab:ablation}, structured reasoning and recovery supervision contribute to different aspects of navigation. Without CoT reasoning, DF increases from 1.85 to 4.30, indicating less stable navigation decisions. In contrast, training only on expert trajectories reduces CSR from 98.15\% to 15.46\%, despite yielding the lowest DF. This result shows that a lower deviation frequency does not imply stronger recovery and highlights the importance of recovery supervision once deviations occur. Coarse and fine grids reduce SR to 48.12\% and 26.78\%, respectively, suggesting that the default better balances spatial precision and grounding complexity. The directional-instruction robustness experiments further distinguish high-level route guidance from local recovery. With 5\% corrupted instructions, SR decreases to 51.50\%, whereas CSR remains 97.62\%. With a fourfold longer update interval, DA-Nav achieves an SR of 50.20\% and a CSR of 97.91\%.

\vspace{-0.3cm}
\subsection{Real-World Deployment}
\label{subsec:real_world}

\subsubsection{System Overview and Instruction Interface}

\begin{figure}[htbp]
    \centering
    \includegraphics[width=0.75\columnwidth]{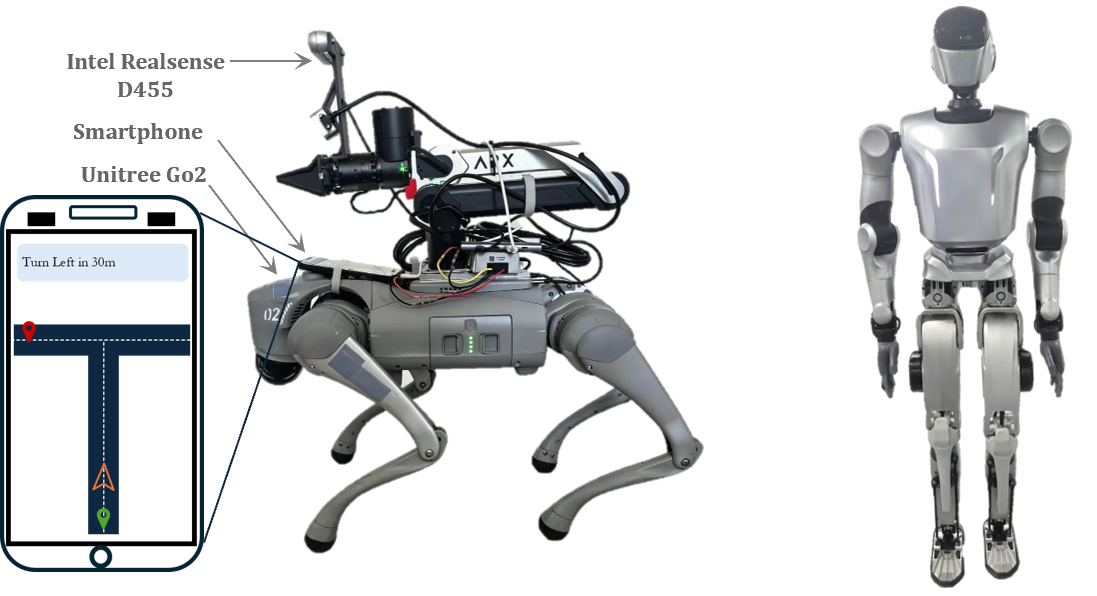}
    \vspace{-0.3cm}
    \caption{Robotic platforms used for real-world deployment. \textbf{Left:} Unitree Go2 quadruped robot. \textbf{Right:} Leju Kuavo-V humanoid robot. Both platforms stream egocentric RGB observations and direction-aware instructions to a remote GPU server for policy inference.}
    \label{fig:hardware}
    \vspace{-0.4cm}
\end{figure}

To validate the zero-shot sim-to-real transferability of DA-Nav, we deploy the framework on robotic platforms, including a Unitree Go2 quadruped and a Leju Kuavo-V humanoid robot, as shown in Fig.~\ref{fig:hardware}. The robots are equipped with a forward-facing Intel RealSense D455 RGB camera for egocentric visual perception. Additionally, each platform is tethered to a smartphone to interface with commercial navigation tools and retrieve real-time direction-aware instructions.

Due to the substantial computational overhead of VLMs, policy inference is executed on a remote GPU server equipped with an NVIDIA RTX 4090. The onboard platform continuously streams multimodal observations and receives structured navigation decisions through a low-latency communication interface for real-time closed-loop execution. To connect coarse-grained commercial navigation tools with local robotic control, we further develop a lightweight navigation instruction parsing module. As illustrated in the inset of Fig.~\ref{fig:hardware}, the module asynchronously extracts real-time path guidance and textual navigation information (e.g., ``Turn left in 30 meters'') from commercial navigation tools and continuously converts them into the structured navigation instruction $I_t$ used by DA-Nav as policy input.

\subsubsection{Open-Loop Primitive Evaluation}

To isolate the execution capability of low-level navigation behaviors without the influence of accumulated errors, we first evaluate open-loop performance on basic primitives. We conduct experiments across 36 different intersections (12 cases each for straight, left, and right turns), with each case evaluated once.

\begin{figure}[htbp]
  \centering
  \includegraphics[width=0.8\columnwidth]{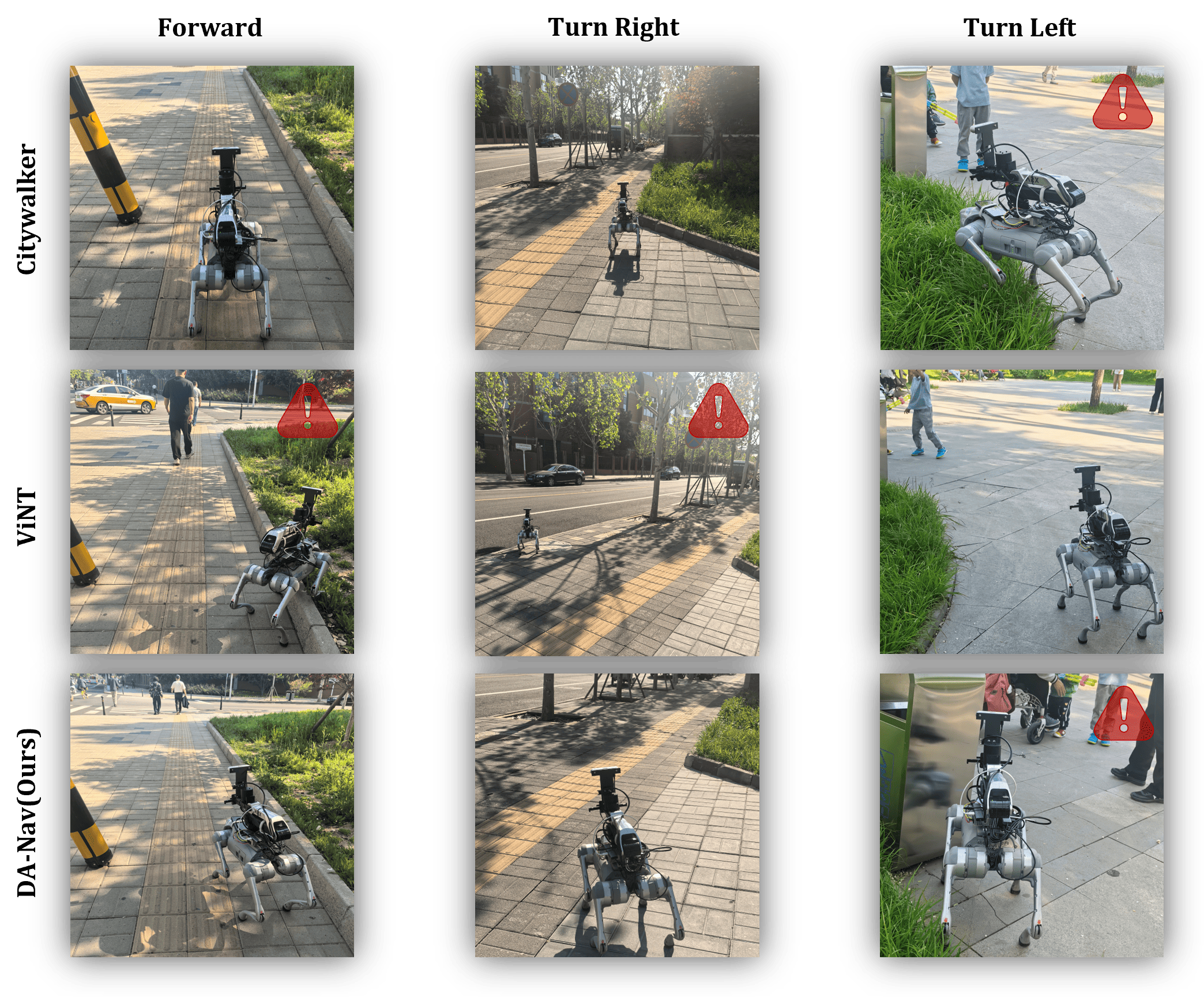}
  \vspace{-0.2cm}
  \caption{Real-world open-loop evaluation of navigation primitives (Forward, Turn Right, and Turn Left) across different methods. Red icons indicate failures or unsafe maneuvers. Compared with CityWalker and ViNT, DA-Nav demonstrates more stable execution and improved spatial grounding during large turning maneuvers.}
\label{fig:open_loop_comparison}
\end{figure}

As reported in Table~\ref{tab:real_world_open} and visualized in Fig.~\ref{fig:open_loop_comparison}, baseline methods demonstrate competitive performance for straight-line navigation, but struggle severely with the large heading changes required for turning. This indicates that large heading changes pose substantial challenges for end-to-end regression-based approaches, which often lack explicit geometric constraints for stable motion generation.

In contrast, by combining structured decision prediction with image-space spatial grounding, DA-Nav demonstrates more stable and geometrically consistent turning behavior, achieving an average success rate of 83.3\%.

\begin{table}[t]
\caption{Real-World Open-Loop Execution of Navigational Primitives}
\vspace{-0.3cm}
\label{tab:real_world_open}
\centering
\resizebox{\columnwidth}{!}{
\begin{tabular}{c ccc c}
\toprule
\textbf{Method} & \textbf{Forward (12)} & \textbf{Turn Left (12)} & \textbf{Turn Right (12)} & \textbf{Avg SR (\%)} \\
\midrule
CityWalker~\cite{CityWalker}  & 100.0 & 66.7 & 58.3 & 75 \\
ViNT~\cite{ViNT}  & 58.3 & 50.0 & 66.7 & 58.3 \\
\midrule
\textbf{DA-Nav (Ours)} & \textbf{100.0} & \textbf{75.0} & \textbf{75.0} & \textbf{83.3} \\
\bottomrule
\end{tabular}
}
\vspace{-0.7cm}
\end{table}

\subsubsection{Closed-Loop Navigation in Real-World Environments}

Following the primitive validation, we evaluate full closed-loop navigation performance across two distinct real-world scenarios: urban streets and parks. For each scenario, five paths are selected (each exceeding 100m with at least one turn), and each path is repeated three times. As reported in Table~\ref{tab:real_world_closed} and visualized in Fig.~\ref{fig:real_world_deploy}, baseline methods perform poorly in real-world closed-loop settings despite having access to pre-recorded path priors. Environmental uncertainties, such as dynamic pedestrians and illumination variations, significantly degrade their performance, resulting in overall success rates of only 23.3\% for CityWalker and 16.7\% for ViNT. In contrast, DA-Nav achieves an overall success rate of 46.7\%. By leveraging direction-aware reasoning and explicit recovery capability, DA-Nav continuously corrects trajectory deviations during execution, ensuring robust navigation under real-world uncertainties. These results highlight that reliable long-horizon navigation critically depends on direction-aware guidance and active recovery capabilities, rather than solely on passive trajectory imitation.

\begin{table}[t]
\caption{Real-World Closed-Loop Navigation Performance}
\vspace{-0.3cm}
\label{tab:real_world_closed}
\centering
\resizebox{\columnwidth}{!}{
\begin{tabular}{c cc cc cc}
\toprule
\multirow{2}{*}{\textbf{Method}} & \multicolumn{2}{c}{\textbf{Urban (15 Trials)}} & \multicolumn{2}{c}{\textbf{Park (15 Trials)}} & \multicolumn{2}{c}{\textbf{Overall}} \\
\cmidrule(lr){2-3} \cmidrule(lr){4-5} \cmidrule(lr){6-7}
& \textbf{SR (\%)} & \textbf{RC (\%)} & \textbf{SR (\%)} & \textbf{RC (\%)} & \textbf{SR (\%)} & \textbf{RC (\%)} \\
\midrule
CityWalker~\cite{CityWalker}  & 20.0 & 47.6 & 26.7 & 58.9 & 23.3 & 53.3 \\
ViNT~\cite{ViNT}  & 13.3 & 43.4 & 20.0 & 54.5 & 16.7 & 49.0 \\
\midrule
\textbf{DA-Nav (Ours)} & \textbf{40.0} & \textbf{68.4} & \textbf{53.3} & \textbf{84.2} & \textbf{46.7} & \textbf{76.3} \\
\bottomrule
\end{tabular}
}
\vspace{-0.4cm}
\end{table}

\begin{figure}[htbp]
    \centering
    \includegraphics[width=0.75\columnwidth]{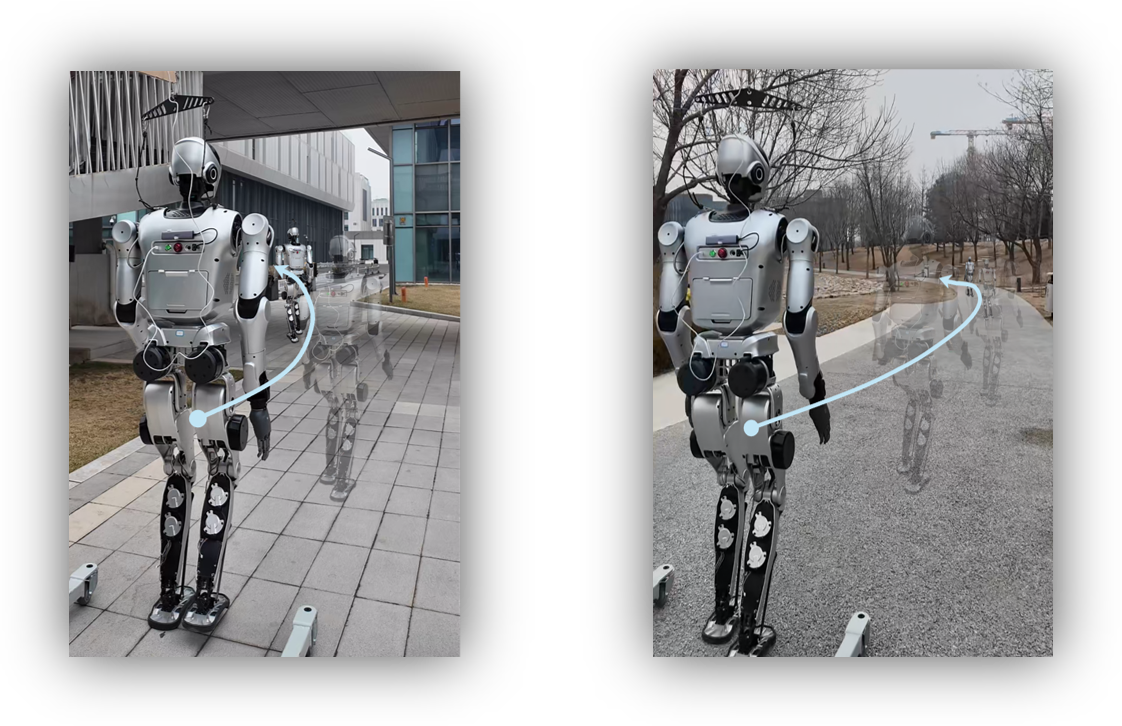}
    \vspace{-0.6cm}
    \caption{Zero-shot cross-embodiment deployment on the Leju Kuavo-V humanoid. The model achieves continuous closed-loop navigation over 1.2 km without real-world fine-tuning, exhibiting stable trajectory execution (blue curves) in both (Left) urban and (Right) park environments.}
    \label{fig:humanoid_realworld}
    \vspace{-0.2cm}
\end{figure}

\subsubsection{Cross-Embodiment Generalization}
Beyond the quadruped platform, we deploy the pre-trained DA-Nav policy without modification onto a full-scale humanoid robot (Leju Kuavo-V), as shown in Fig.~\ref{fig:humanoid_realworld}. Without embodiment-specific fine-tuning, it achieves successful goal-directed navigation over 1.2 km in outdoor environments, demonstrating strong cross-embodiment generalization.

\vspace{-0.15cm}
\section{Conclusion}
\label{sec:conclusion}

In this paper, we present DA-Nav, a direction-aware framework that integrates CoT reasoning for trajectory recovery in robust long-horizon closed-loop navigation. It achieves stable navigation under both simulated and real-world environmental perturbations. Extensive experiments in CARLA and real-world deployments across quadruped and humanoid platforms demonstrate strong generalization, autonomous recovery capability, and cross-embodiment transferability.

Despite these advantages, the current system still relies on commercial navigation tools for high-level path guidance, which introduces practical limitations such as GPS inaccuracies, update latency, and restricted coverage in unmapped regions. Future work will focus on eliminating this dependency through autonomous exploration and lifelong topological memory construction. By enabling robots to continuously build persistent spatial representations in unknown environments, we aim to achieve robust long-horizon navigation in fully unstructured and GPS-denied scenarios.









\end{document}